\def\year{2022}\relax
\documentclass[letterpaper]{article} 
\usepackage{aaai22} 
\usepackage{times} 
\usepackage{helvet} 
\usepackage{courier} 
\usepackage[hyphens]{url} 
\usepackage{graphicx} 
\usepackage{subfigure}
\usepackage{multirow}
\usepackage{natbib} 
\usepackage{caption} 
\DeclareCaptionStyle{ruled}{labelfont=normalfont,labelsep=colon,strut=off} 
\usepackage{algorithm}
\usepackage{algorithmic}

\usepackage{amsmath} 
\usepackage{amssymb} 
\usepackage{newfloat}
\usepackage{listings}
\floatstyle{ruled}
\newfloat{listing}{tb}{lst}{}
\floatname{listing}{Listing}
\usepackage{fancyhdr}

\title{Multi-task Self-distillation for Graph-based Semi-Supervised Learning}

\author {
	Yating Ren\textsuperscript{\rm a}, 
	Junzhong Ji\textsuperscript{\rm b}, 
	Lingfeng Niu\textsuperscript{\rm c}, 
	Minglong Lei\textsuperscript{\rm d,\thanks{leiml@bjut.edu.cn}} 
	\\
}
\affiliations {
	\textsuperscript{\rm a,b,d} Beijing University of Technology, \textsuperscript{\rm c} University of the Chinese Academy of Sciences.
}

\begin{document}
	
	\maketitle
	\thispagestyle{fancy}
	\fancyhead{}
	\lhead{This work has been submitted to the IEEE for possible publication. Copyright may be transferred without notice, after which this version may no longer be accessible.}
	\lfoot{}
	\cfoot{}
	\rfoot{}
	
	\begin{abstract}
		Graph convolutional networks have made great progress in graph-based semi-supervised learning. Existing methods mainly assume that nodes connected by graph edges are prone to have similar attributes and labels, so that the features smoothed by local graph structures can reveal the class similarities. However, there often exist mismatches between graph structures and labels in many real-world scenarios, where the structures may propagate misleading features or labels that eventually affect the model performance. In this paper, we propose a multi-task self-distillation framework that injects self-supervised learning and self-distillation into graph convolutional networks to separately address the mismatch problem from the structure side and the label side. First, we formulate a self-supervision pipeline based on pre-text tasks to capture different levels of similarities in graphs. The feature extraction process is encouraged to capture more complex proximity by jointly optimizing the pre-text task and the target task. Consequently, the local feature aggregations are improved from the structure side. Second, self-distillation uses soft labels of the model itself as additional supervision, which has similar effects as label smoothing. The knowledge from the classification pipeline and the self-supervision pipeline is collectively distilled to improve the generalization ability of the model from the label side. Experiment results show that the proposed method obtains remarkable performance gains under several classic graph convolutional architectures.
	\end{abstract}
	
	\section{Introduction}
	Semi-supervised learning on graphs (GSSL) is a fundamental machine learning task with only limited labels for graph nodes available. The goal of GSSL is to leverage the small proportion of labeled nodes in conjunction with abundant graph structures to classify the rest of unlabeled nodes~\cite{zhu2005semiGraph}. Successfully resolving GSSL problems provides support for many downstream applications, e.g., POI recommendations~\cite{yang2017bridging}, hyperspectral image classification~\cite{shao2018spatial}, phenotype classification~\cite{doostparast2018graph} and part-of-speech tagging~\cite{subramanya2010efficient}.
	
	Recently, graph convolutional networks (GCNs) show great potentials in GSSL problems~\cite{kipf2016semiGCN}, which mainly benefits from the local smoothing operations that aggregate attributes from neighbors to generate discriminative features~\cite{li2018deeper}. Many followed GCN models devote efforts to developing efficient aggregation functions to learn more powerful features from graph structures~\cite{velivckovic2017GAT, Klicpera2019APPNP}. In general, the success of GSSL often requires efforts from both graph structures and labels. The fractional labels can be combined with graph structures to provide additional supervision. For example, label propagation algorithm (LPA) uses label smoothing to match the attributes and labels~\cite{li2019label, wang2020unifying}. Within the GCN framework, they proved that the feature aggregation gives theoretical guarantees for label propagation.
	
	Notice that the combination of graph structures and labels only works effectively when the edges in graphs reveal the real feature or label similarities. Unfortunately, this assumption is often violated in real-world scenarios since there exist mismatches between graph structures and labels~\cite{Yuto2017LPfail, chien2020adaptive}. For example, in a citation network, a paper may cite papers from other fields. In other words, simply exploring the proximity exhibited by edges is insufficient to discover qualified levels of node similarities for inferring the labels. Besides, many methods that augment the limited labels by graphs are usually based on hard labels, where the distributional information that reveals the label similarities cannot be adequately captured during the training.
	
	In this paper, we address these challenges by whittling down the mismatches between graph structures and label similarities, instead of directly using graph connections as vehicles to propagate hard label information. We cast our multi-task self-distillation framework by integrating graph-based self-supervised learning and self-distillation into GCNs (SDSS-GCN). From the \emph{structure side}, graph-based self-supervised learning mines underlying structural similarities and is able to learn general features that facilitate semi-supervised learning. To this end, we propose to use different levels of similarities (e.g., node level, community level, and graph level) to design pre-text tasks, so that the proximity that may contribute to the prediction can be fully explored. From the \emph{label side}, the soft predictions of the model itself are leveraged to capture the label distributions. Aside from the ability to improve model generalization, self-distillation is also closely related to the label smoothing mechanism and is working as regularization for GCNs~\cite{zhang2020self}.
	
	To be concrete, we built a two-stage training architecture where self-distillation is implemented based on middle-layer outputs, classification outputs and self-supervision outputs. By jointly optimizing the self-supervision pipeline and the classification pipeline, the knowledge encapsulated in graph structures and labels is carefully explored. It is delightful to find that self-supervised learning and self-distillation are nicely assembled under the GCN framework. First, incorporating self-supervised learning may bring undesired guidance that affects the model training. Self-distillation can improve the stability of training in a teacher-free fashion. Second, the distillation of self-supervision outputs considers more information that can further improve the generalization of GCNs. The contributions of our model are summarized in four-fold:
	\begin{itemize}
		\item We propose a novel self-distilled multi-task GCN framework named SDSS-GCN for semi-supervised learning, which further mines the information within graphs and labels to resolve the mismatches between node proximity and label similarities.
		\item We resort to self-supervised learning based on four pre-text tasks to extract different levels of proximity. The improved feature extraction process largely facilitates the local aggregation in GCNs.
		\item We propose to use self-distillation that is highly related to label smoothing to further improve the generalization of GCNs. The soft labels of the model itself provide distributional label information that can be matched with the structures more easily.
		\item Extensive experiments show that self-supervision and self-distillation are nicely incorporated in the GCN framework, and achieves impressive performance gains in several widely-used GCN-based frameworks.
	\end{itemize}

	\section{Background and Related Works}\label{sec:section2}
	
	\paragraph{Graphs.}\label{sec:section2.1} We use $\mathcal{G}=(\mathcal{V}, \mathcal{E})$ to denote a graph where $\mathcal{V}=\left\{v_{1}, v_{2}, \ldots, v_{n}\right\}$ is the set of $n$ nodes, $\mathcal{E}$ is the set of edges describing the relations between nodes. The graph structure information can also be represented by an adjacency matrix $\mathbf{A} \in[0,1]^{n \times n}$ where $\mathbf{A}_{i,j}=1$ indicates that there exists a edge between nodes $v_{i}$ and $v_{j}$, otherwise $\mathbf{A}_{i,j}=0$. The feature matrix for all nodes is denoted as $\mathbf{X}$, where $\mathbf{x}_i$ is the feature vector of node $v_i$.
	
	\paragraph{Graph-based Semi-supervised Learning.}
	In this paper, we focus on the semi-supervised node classification task where only a subset of nodes $\mathcal{V}_{L} \subset \mathcal{V}$ with $|\mathcal{V}_{L}|\ll|\mathcal{V}|$ are associated with labels drawn from a label set $L = \{1, \cdots, C \}$. We can also denote the labels of all nodes as a label matrix $\mathbf{Y}$, where $\mathbf{y}_i$ is the one-hot label vector for node $v_i$. The graph-based semi-supervised learning aims at taking advantage of the graph $\mathcal{G}$, node features $\mathbf{X}_{1:|V_L|}$, and node labels $\mathbf{Y}_{1:|V_L|}$ to train a classifier that can infer the labels of nodes in unlabeled node set $\mathcal{V}_{U} \subset \mathcal{V}$  ($\mathcal{V}_{U} = \mathcal{V}\setminus \mathcal{V}_{L}$). The objective to be optimized is the differences between the predictions and ground truth labels. Formally, the loss function can be represented as, 
	\begin{equation}
		\begin{aligned}
			\mathcal{L}_{\textit{NC}}=\frac{1}{|\mathcal{V}_{L}|}\sum_{v_{i}\in \mathcal{V}_{L}}\operatorname{dist}\left(
			F(\mathbf{A},\mathbf{X},\mathcal{N}(v_i)), \mathbf{y}_{i}
			\right),
		\end{aligned}\label{eq:goal}
	\end{equation}
	where $\mathcal{N}(v_i)$ denotes the neighbors of $v_i$, $F$ is the mapping function from the input to the predictions, $\operatorname{dist}(\cdot,\cdot)$ is the distance function that measures the differences (e.g., cross entropy), and $\mathcal{L}_{\textit{NC}}$ is the loss function for node classification. Traditional GSSL approaches are mainly based on graph regularizations~\cite{zhou2004learning, belkin2006manifold}. Later approaches are mostly based on graph embedding methods~\cite{Yang2016Revisiting} that spur the development of advanced graph neural networks~\cite{kipf2016semiGCN, Hamilton2017GraphSAGE, velivckovic2017GAT, feng2020grand, li2020co}. 
	
	\paragraph{Graph Convolutional Networks.} GCN is a multi-layer neural network that iteratively aggregates features through the edges. The utilizing of local information makes it effective in graph-based semi-supervised learning. The vanilla GCN~\cite{kipf2016semiGCN} is a two-layer neural network that can be formulated as, 
	\begin{equation}
		\begin{aligned}
			\mathbf{Z}=\mathbf{L}\left(\operatorname{ReLU}({\mathbf{L}}\mathbf{X}\mathbf{W}^{(0)})\right)\mathbf{W}^{(1)},
		\end{aligned}\label{eq:gcn}
	\end{equation}
	where $\mathbf{Z}$ is the output logits matrix, $\mathbf{L}=\hat{\mathbf{D}}^{-\frac{1}{2}} \hat{\mathbf{A}}\hat{\mathbf{D}}^{-\frac{1}{2}}$, $\hat{\mathbf{A}}$ is the adjacent matrix with self-loop, $\hat{\mathbf{D}}$ is the degree matrix of $\hat{\mathbf{A}}$, $\mathbf{W}^{(0)}$ and $\mathbf{W}^{(1)}$ are parameter matrices. 
	
	The forward process of GCN in Eqn.~\eqref{eq:gcn} can be generalized as the combination of a feature propagation process and a feature transformation process between $l$-th and $(l+1)$-th layer~\cite{wu2019SGCNs}. The feature propagation is to smooth the node features by the adjacent matrix: $\mathbf{L}\mathbf{H}^{(l)} \rightarrow \mathbf{H}^{(l+\frac{1}{2})}$ while the feature transformation is to transform the features via parameter matrix $\mathbf{W}^{(l)}$: $\operatorname{ReLU}(\mathbf{H}^{(l+\frac{1}{2})} \mathbf{W}^{(l)}) \rightarrow \mathbf{H}^{(l+1)}$. We use $\mathbf{H}^{(l+\frac{1}{2})}$ to denote the intermediate results after feature propagation.
	
	Let $\phi(\mathbf{X},{\mathbf{A}})=\mathbf{L}\left(\operatorname{ReLU}({\mathbf{L}}\mathbf{X}\mathbf{W}^{(0)})\right)$ in Eqn.\eqref{eq:gcn} be the feature extractor of GCN parameterized with $\mathbf{\Omega}$. The graph-based semi-supervised learning based on GCN can be decomposed into a feature extraction function $\phi(\cdot)$ and a linear transformer $\mathbf{\Theta}$: $\mathbf{Z}=\phi(\mathbf{X},\mathbf{A})\mathbf{\Theta}$, where $\mathbf{\Theta}=\mathbf{W}^{(1)}$. Thus, Eqn.~\eqref{eq:goal} can be crystallized as,
	\begin{equation}
		\begin{aligned}
			\mathcal{L}_{\textit{NC}}=\frac{1}{|\mathcal{V}_{L}|}\sum_{v_{i}\in \mathcal{V}_{L}}\operatorname{dist}(\mathbf{z}_{i},\mathbf{y}_{i}),
		\end{aligned}\label{eq:gcnloss}
	\end{equation}
	where $\mathbf{z}_{i}$ is the output logits of node $v_i$.
	
	\section{Method}
	
	\begin{figure*}[htb]
		\centering
		\includegraphics[width=1\textwidth]{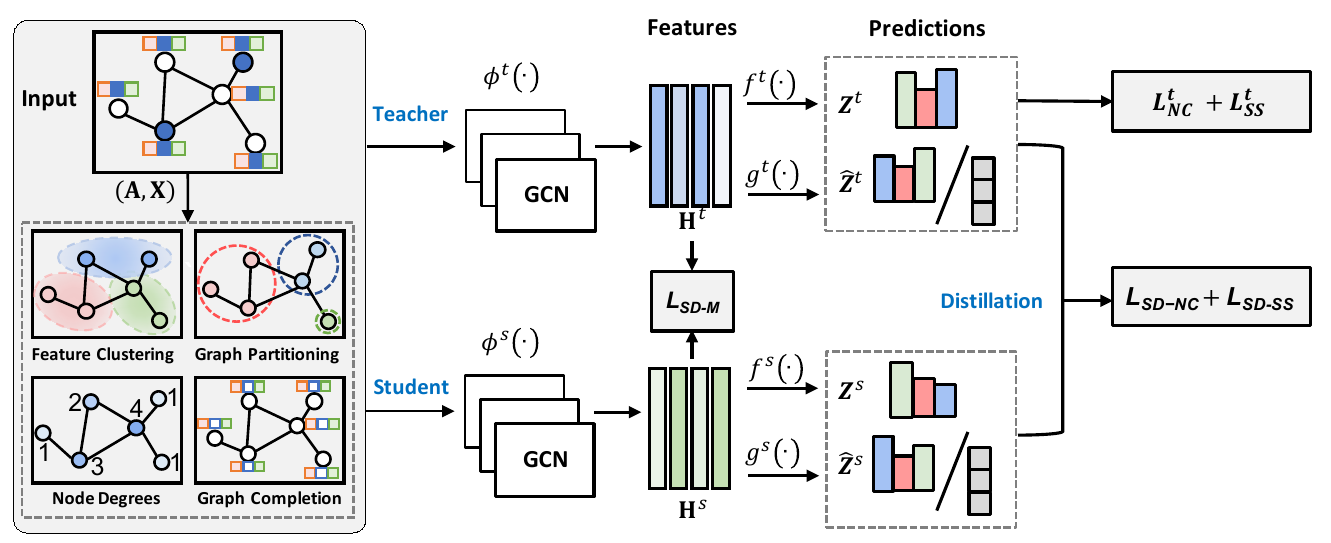}
		\caption{The overview of our framework. The left side gives the process of building the self-supervision tasks. The right side describes the self-distillation process where the teacher network (top) and the student network (bottom) share the same structures. The teacher and student network both consist of a classification pipeline and a self-supervision pipeline, where the same feature extraction backbone is used for the two pipelines. The distillation is accomplished from the classification outputs, self-supervision outputs, and middle-layer outputs.}
		\label{fig:framework}
	\end{figure*}
	
	To resolve the mismatch problem between structures and labels, self-supervised learning and self-distillation strategies are used to mine the information from both structures and labels. The overall architecture is illustrated in Figure~\ref{fig:framework}. Notice that our framework distills knowledge from the model itself in a teacher-free fashion and is more compatible with current graph convolutional models. Different levels of proximity are plugged into the structural information extraction backbone. The details of our framework are described in the following subsections.
	
	\subsection{Graph-based Self-supervised Learning}
	Even graph convolutional networks are powerful feature extractors, the inner information of the data has not been fully utilized~\cite{wan2021contrastive}. In this subsection, we build a multi-task framework that leverages self-supervised learning to increase data efficiency by further mining the information of the data itself without any label information. 
	
	At present, self-supervised learning has been widely used in situations with limited labels. We can formulate the general process of self-supervised learning based on GCNs similar to the node classification task,
	\begin{equation}
		\begin{aligned}
			\hat{\mathbf{Z}}=\phi(\hat{\mathbf{X}},\hat{\mathbf{A}})\hat{\mathbf{\Theta
			}},
		\end{aligned}\label{eq:ssz}
	\end{equation}
	where $\hat{\mathbf{\Theta}}$ is the linear transformation, $\hat{\mathbf{X}}$ and $\hat{\mathbf{A}}$  are the inputs of pre-text tasks, $\hat{\mathbf{Z}}$ is the self-supervision predictions. The corresponding self-supervision loss function $\mathcal{L}_{\textit{SS}}$ can be formulated as,
	\begin{equation}
		\begin{aligned}
			\mathcal{L}_{\textit{SS}}=\frac{1}{|\mathcal{V}_{L}|}\sum_{v_{i}\in \mathcal{V}_{L}}\operatorname{dist}(\hat{\mathbf{z}}_{i},\hat{\mathbf{y}}_{i}),
		\end{aligned}\label{eq:ssloss}
	\end{equation}
	where $\hat{\mathbf{z}}_{i}$ is the predicted logits for pre-text tasks, $\hat{\mathbf{y}}_{i}$ is the ground truth of $v_{i}$ from $\hat{\mathbf{Y}}$, and $\operatorname{dist}(\cdot,\cdot)$ is the distance function. We use the features of $\mathcal{V}_{L}$ during the training under the inductive setting. Different from the target task, self-supervised learning contains both classification tasks and regression tasks. We use the cross entropy as distance function for classification tasks and use smooth mean absolute error for regression tasks. Concretely, the smooth mean absolute error can be formulated as,
	\begin{equation}
		\begin{aligned}
			\operatorname{dist}(\hat{\mathbf{z}},\hat{\mathbf{y}})=
			\begin{cases}
				0.5*(\hat{\mathbf{z}}-\hat{\mathbf{y}})^{2}, &\text{if } |\hat{\mathbf{z}}-\hat{\mathbf{y}}|<1, \\
				|\hat{\mathbf{z}}-\hat{\mathbf{y}}|-0.5. & \text{otherwise.} 
			\end{cases}
		\end{aligned}\label{eq:smoothL1}
	\end{equation}
	
	The self-supervised task can be co-trained with the target task to formulate a multi-task framework. We jointly optimize the following loss function,
	\begin{equation}
		\begin{aligned}
			\mathcal{L} = \mathcal{L}_{\textit{NC}}+\alpha \mathcal{L}_{\textit{SS}},
		\end{aligned}\label{eq:gcnssloss}
	\end{equation}
	where $\alpha$ is a positive hyper-parameter. The multi-task training paradigm largely facilitates the feature extraction process of GCNs since it introduces knowledge from the data itself that can improve the generalization of the features~\cite{Yuning2020Whendoes, Zhu2020MultiStageSS}. To capture different levels of graph similarities, we design several pre-text tasks for GCNs based on graph properties and prior knowledge.
	
	\paragraph{Node Degrees.} Node degree is an essential graph property where its distribution characterizes the network influence. Many phenomena in graphs, e.g., random walks and diffusion, are highly related to node degrees. Hence, we build the self-supervision task that predicts the node degrees based on node features and adjacent matrix. Notice that this task is unscathed since it requires no pre-processing procedure to the inputs. Consequently, the pre-text task and the target task have the same inputs during the training. Formally, we denote the continuous node degrees as labels $\mathbf{d} \in \mathbb{R}^{n}$ where $d_{i} = \sum_{j} \mathbf{A}_{i,j}$. Then, predicting the node degrees is formulated as a regression problem that minimizes the distance between the predicted degrees $\phi(\mathbf{A},\mathbf{X})\hat{\mathbf{\Theta}}$ and ground truth $\mathbf{d}$.
	
	\paragraph{Feature Clustering.} Clustering is a classical unsupervised task that can capture the community-wise proximity in graphs. There are several attempts that use clustering to build pre-text tasks for self-supervised learning~\cite{Zhu2020MultiStageSS, Yuning2020Whendoes}. Grouping nodes into dense clusters detects attributed similarities that are beneficial for classifications. To construct the pre-text task, we firstly run $k$-means on the input attributes and assign each node to a cluster $C_{i}$ where $i \in \{ 1,2, \cdots, K\}$ and $K$ is the total number of clusters. Then, we can take the clustering assignments as labels and feed them into GCNs to assist the training. Notice that instead of using the features calculated by GCNs, we emphasize the role of initial attributes that have not been smoothed by structures to provide the complete attribute-perspective information.  
	
	\paragraph{Graph Partitioning.} Graph partitioning is another community-wise pre-text task based on graph typologies. It aims to divide nodes into different subsets where inter-connections are sparse and intra-connections are dense. Unlike feature clustering, graph partitioning captures structural similarities only provided by connections. Additionally, graph partitioning considers global information, which is complementary for GCNs that focus on local information aggregation. To be concrete, we select METIS~\cite{Karypis1998Partitioning} to accomplish this goal. Given a graph $\mathcal{G}$, METIS generates $K$ distinct node sets $C_{i}$ where $i \in \{ 1,2, \cdots, K\}$. To avoid extremely unbalanced partitioning, the sizes of partitioned sets are constrained by $K\frac{\max_{i} | C_{i} |}{|\mathcal{V}|} \leq 1+\epsilon$ where $\epsilon$ is a small value between 0 and 1. Finally, the partitioning results are served as labels to guide the training of the self-supervised learning pipeline. 
	
	\paragraph{Graph Completion.} Predicting the missing attributes in graphs brings deep understandings of the local smoothness in graphs. From the perspective of graph signal processing, the local continuity describes how signals distribute along with graph structures. Such an assumption also spurs the Laplacian regularizations in graph theory. Consequently, we build a pre-text task that aims to predict the attributes of missing nodes. The features of randomly selected nodes are firstly masked. Then, the goal turns to predict the unscathed graphs by the masked graphs. We notice that the attributes in graphs are high-dimensional, which brings additional computational burdens for the multi-task framework. To solve this problem, we employ Principle Component Analysis (PCA) to reduce the dimension of features. The key advantage of applying dimension reduction to attributes is that the self-supervised learning component is encouraged to focus on the salient information in the attributes.

	\subsection{Graph-based Self-Distillation}
	After introducing the self-supervision, we harness the self-distillation strategy into GCNs to distill knowledge from the multi-task framework. Self-distillation can transfer the knowledge from the model itself in a teacher-free fashion. In other words, the teacher network and the student network share the same structure. The knowledge from the ``future'' can stabilize the training process. Unlike previous knowledge distillation frameworks that only distill knowledge from the target task, we utilize supervisions from both the target and auxiliary tasks. As indicated in Figure~\ref{fig:framework}, both the teacher and the student consist of three components: a feature extractor $\phi(\cdot)$ with parameter $\mathbf{\Omega}$, a linear transformation layer $f(\cdot)$ with parameter $\mathbf{\Theta}$ for the node classification, and a linear transformation layer $g(\cdot)$ with parameter $\hat{\mathbf{\Theta}}$ for the pre-text task.
	
	The incorporating of self-distillation constitutes a two-stage training scheme. In the first stage, the node classification and pre-text tasks are trained until its convergence by optimizing the objective in Eqn. \ref{eq:gcnssloss}. We can obtain the soft logits $\mathbf{Z}^{t}$ for the node classification and the logits $\hat{\mathbf{Z}}^{t}$ for the self-supervised learning . For the target task, the soft logits carry the distributional information of labels that are more adequate measurements for label similarities compared with hard labels~\cite{hinton2015distilling}.
	
	In the second stage, we train the student network with the supervision of the teacher. The knowledge of the multi-task framework is transferred to the student by two self-distillation losses, $\mathcal{L}_{\textit{SD-NC}}$ and $\mathcal{L}_{\textit{SD-SS}}$, where the former is for the node classification while the latter is for pre-text tasks. Still, in most GCNs, the feature forward process in neural networks and the feature aggregation in graphs are highly intertwined. In other words, one forward GCN layer propagates features only from one-hop neighbors. This phenomenon limits the depths of many GCN-based models, which makes the information in each immediate layer important. We further propose to distill knowledge from the intermediate layer of GCNs by optimizing the loss $\mathcal{L}_{\textit{SD-M}}$. 
	
	To sum up, the student network is trained with respect to the follow loss,
	\begin{equation}
		\begin{aligned}
			\mathcal{L}_{\textit{SD}}=\mathcal{L}_{\textit{SD-NC}} + \mathcal{L}_{\textit{SD-SS}} +  \mathcal{L}_{\textit{SD-M}},
		\end{aligned}\label{eq:sdloss}
	\end{equation}
	where $\mathcal{L}_{\textit{SD}}$ is the overall self-distillation loss. We give the details of three losses as follows.
	
	\paragraph{Distillation for Node Classification.}
	To balance the hard labels and soft labels, the node classification pipeline in the student network is supervised by the combination of two terms. Using KL divergence as the distance measure, the optimization objective can be denoted as,
	\begin{equation}
		\begin{aligned}
			\mathcal{L}_{\textit{SD-NC}} = -\beta_{1} \sum_{\mathbf{p}} \sum_{j=1}^{m} p_{j}^{t} \log p_{j}^{s} -(1-\beta_{1})\sum_{\mathbf{y}'} \sum_{j=1}^{m} y'_{j} \log y_{j}^{s},
		\end{aligned}\label{eq:Lsdnc}
	\end{equation}
	where superscripts $t$ and $s$ denote the teacher and the student, $\mathbf{p}=\operatorname{softmax}(\mathbf{z}/\tau)$, $\tau$ is the temperature, $\mathbf{y}'$ is the ground truth, $m$ is the number of classes, and $\beta_{1}$ is a hyper-parameter that controls the ratio of supervision from the teacher. 
	
	\paragraph{Distillation for Self-supervision.} 
	We can distill the knowledge from self-supervised learning similar to the node classification task. The optimization objective is,
	\begin{equation}
		\begin{aligned}
			\mathcal{L}_{\textit{SD-SS}} = -\beta_{2} \sum_{\hat{\mathbf{p}}} \sum_{j} \hat{p}_{j}^{t} \log \hat{p}_{j}^{s} -(1-\beta_{2})\sum_{\hat{\mathbf{y}}'} \sum_{j} \hat{y}'_{j} \log \hat{y}_{j}^{s},
		\end{aligned}\label{eq:Lsdss}
	\end{equation}
	where $\hat{\mathbf{p}}=\operatorname{softmax}(\hat{\mathbf{z}}/\tau)$, $\hat{\mathbf{y}}'$ is the ground truth, and $\beta_{2}$ is the balance hyper-parameter.
	
	\paragraph{Distillation for Intermediate Layer.}
	To distill knowledge from shallow layers, we fetch the hidden outputs $\mathbf{H}^{t}$ and $\mathbf{H}^{t}$ from the feature extractor. It aligns the middle-layer outputs of the student network via the following function,
	\begin{equation}
		\begin{aligned}
			\mathcal{L}_{\textit{SD-M}} = \sum_{\mathcal{V}_{L}} \operatorname{dist} (\mathbf{H}^{t}, \mathbf{H}^{s}
			),
		\end{aligned}\label{eq:lossdh}
	\end{equation}
	where $\operatorname{dist}(\cdot,\cdot)$ is the smooth mean absolute error in Eqn.~\eqref{eq:smoothL1}.
	
	\subsection{Overall Algorithm}
	The overall algorithm is listed in Algorithm \ref{alg:algorithm}. First, we train the teacher network which includes a node classification loss and a self-supervised learning loss. Second, the student network is trained with the supervision of the distillation from the teacher network. After the two-stage training, the student network is used for the inference.
	
	\begin{algorithm}[!h]
		\caption{Training Process of Our Method}
		\label{alg:algorithm}
		\begin{algorithmic}[1]
			\REQUIRE Node features $\mathbf{X}_{1:|V_L|}$, adjacent matrix $\mathbf{A}$, node labels $\mathbf{Y}_{1:|V_L|}$, hyper-parameters $\alpha$, $\beta_1$ and $\beta_2$.
			\ENSURE Model parameters $\mathbf{\Omega}$, $\mathbf{\Theta}$ and $\hat{\mathbf{\Theta}}$.
			\STATE \textbf{[Training Teacher Network]}
			\FOR {epoch $\leq$ max epochs}
			\STATE Obtain input $\hat{\mathbf{X}}$ for the pre-text task;
			\STATE Obtain middle-layer outputs $\mathbf{H}^{t}$ and $\hat{\mathbf{H}}^{t}$ via $\phi^{t}(\cdot)$;
			\STATE Obtain soft logits $\mathbf{Z}^{t}$ and $\hat{\mathbf{Z}}^{t}$ via $f^{t}(\cdot)$ and $g^{t}(\cdot)$;
			\STATE Calculate the teacher loss via Eqn.~\eqref{eq:gcnssloss};
			\STATE Update parameters $\mathbf{\Omega}^{t}$, $\mathbf{\Theta}^{t}$ and $\hat{\mathbf{\Theta}}^{t}$ for $\phi^{t}$, $f^{t}$ and $g^{t}$;
			\ENDFOR
			\STATE \textbf{[Training Student Network]}
			\FOR {epoch $\leq$ max epochs}
			\STATE Obtain input $\hat{\mathbf{X}}$ for the pre-text task;
			\STATE Obtain middle-layer outputs $\mathbf{H}^{s}$ and $\hat{\mathbf{H}}^{s}$ via $\phi^{s}(\cdot)$;
			\STATE Obtain soft logits $\mathbf{Z}^{s}$ and $\hat{\mathbf{Z}}^{s}$ via $f^{s}(\cdot)$ and $g^{s}(\cdot)$;
			\STATE Calculate the distillation loss via Eqn.~\eqref{eq:sdloss};
			\STATE Update parameters $\mathbf{\Omega}^{s}$, $\mathbf{\Theta}^{s}$ and $\hat{\mathbf{\Theta}}^{s}$ for $\phi^{s}$, $f^{s}$ and $g^{s}$;
			\ENDFOR
			\STATE Return $\mathbf{\Omega}^{s}$ and $\mathbf{\Theta}^{s}$.
		\end{algorithmic}
	\end{algorithm}

	\begin{table*}[htbp]
		\centering
		\begin{tabular}{c|cccc|cccc|cccc}
			\hline
			Dataset 
			& \textbf{GCN} & +SS &+SD & +SDSS & \textbf{GAT} & +SS &+SD & +SDSS & \textbf{SAGE} &+SS &+SD & +SDSS \\ \hline
			Cora 
			& 81.83 &83.09 & 85.10 & \textbf{86.00}& 
			83.89 &83.98 & 83.65 &\textbf{ 85.29} & 
			81.78 &83.19 & 84.91 & \textbf{86.00 } \\
			Citeseer 
			& 69.83 &71.22 &73.70 & \textbf{76.13 } & 
			72.76 &73.26 & 74.03 & \textbf{76.35} & 
			66.96 & 72.60 & 72.76 & \textbf{74.20} \\
			Pubmed 
			& 77.52 &79.86 &79.52 & \textbf{82.21} & 
			77.02 & 78.48 & 79.38 & \textbf{81.46} & 
			77.36 &79.36 & 80.52 & \textbf{82.17} \\
			A-Comp 
			& 83.18 &84.22 &84.56 & \textbf{84.86} & 
			81.07 &83.45 &82.03 & \textbf{84.02} & 
			77.60 &83.97 & 82.74 & \textbf{84.66} \\ \hline
		\end{tabular}
		\caption{Ablation study on GCN, GAT and GraphSAGE.}
		\label{tab:res1}
	\end{table*}

	\begin{table}[htbp]
		\centering
		\begin{tabular}{lcccc}
			\hline
			Method & Cora & Citeseer & Pubmed & A-Comp \\ \hline
			GCN & 81.83 & 69.83 & 77.52 & 83.18 \\
			GAT & 83.89 & 72.76 & 77.02 & 81.07 \\
			GraphSAGE & 81.78 & 66.96 & 77.36 & 77.60 \\
			GCNII & 83.69 & 72.76 & 78.02 & 83.04 \\
			APPNP & 83.75 & 71.60 & 79.50 & 81.76 \\
			SGC & 80.52 & 71.33 & 78.92 & 82.48 \\
			DropEdge & 82.80 & 72.30 & 79.60 & 83.40 \\
			FastGCN & 81.90 & 69.70 & 78.10 & 82.10 \\
			MixHop & 82.30 & 72.20 & 81.40 & 80.90 \\
			Graph U-Net & 85.00 & 73.70 & 79.80 & 83.10 \\
			GMNN & 83.70 & 72.90 & 81.80 & 83.30 \\
			GRAND & 85.80 & 75.80 & 82.70 & 84.70 \\\hline
			SDSS-GCN &\textbf{ 86.00} & 76.13 & 82.21 & \textbf{84.86} \\ 
			SDSS-GAT & 85.29 & \textbf{76.35} & 81.46 & 84.02 \\
			SDSS-SAGE & \textbf{86.00} & 74.20 & 82.17 & 84.66 \\
			SDSS-GCNII & 85.85 & 74.75 & 81.82 & 84.07 \\
			SDSS-APPNP & 85.90 & 75.75 & \textbf{82.72} & 84.66 \\ \hline
		\end{tabular}
		\caption{Overall comparison with the state-of-the-arts.}
		\label{tab:Sota}
	\end{table}
	
	\begin{table*}[htbp]
		\centering
		\begin{tabular}{c|cccc|cccc|cccc}
			\hline
			\multirow{2}{*}{Dataset} & \multicolumn{4}{c|}{SDSS-GCN} & \multicolumn{4}{c|}{SDSS-GAT} & \multicolumn{4}{c}{SDSS-SAGE} \\ \cline{2-13} 
			& Deg. & Clu. & Part. & \multicolumn{1}{c|}{Comp.} & Deg. & Clu. & Part. & \multicolumn{1}{c|}{Comp.} & Deg. & Clu. & Part. & Comp. \\ \hline
			Cora & 85.53 & \textbf{86.00} & 83.51 & 85.95 & 85.15 & 84.96 & \textbf{85.29} & 84.87 & \textbf{86.00} & 85.43 & 84.26 & 84.22 \\
			Citeseer & \textbf{76.13} & 72.98 & 72.10 & 73.37 & \textbf{76.35} & 75.69 & 75.41 & 75.36 & 70.39 & 70.00 & \textbf{74.20} & 71.10 \\
			Pubmed & 80.23 & 77.85 & 80.96 & \textbf{82.21} & 80.41 & 80.51 & \textbf{81.46} & 78.87 & \textbf{82.01} & 81.11 & 80.97 & 81.96 \\
			A-Comp & 84.43 & \textbf{84.86} & 84.46 & 84.04 & 83.24 & \textbf{84.24} & 84.10 & 84.02 & 84.18 & 83.92 & 84.19 & \textbf{84.66} \\ \hline
		\end{tabular}
		\caption{Comparison under four different auxiliary tasks.}
		\label{tab:task1}
	\end{table*}
	
	\begin{table*}[htbp]
		\centering
		\begin{tabular}{c|ccc|ccc|ccc}
			\hline
			\multirow{2}{*}{Dataset} & \multicolumn{3}{c|}{GCN} & \multicolumn{3}{c|}{GAT} & \multicolumn{3}{c}{SAGE} \\ 
			\cline{2-10} 
			& NC & NC+M & NC+SS+M & NC & NC+M & NC+SS+M & NC & NC+M & NC+SS+M \\
			\hline
			Cora & 84.22 & 85.10 & \textbf{86.00} & 83.56 & 83.56 & \textbf{85.29} & 83.57& 83.89 & \textbf{86.00} \\
			Citeseer & 71.93 & 73.70 & \textbf{76.13} & 74.09 & 74.14 & \textbf{76.35} & 70.00 & 70.06 & \textbf{74.20} \\
			Pubmed & 75.88 & 77.52 & \textbf{82.21} & 79.26 & 79.28 & \textbf{81.46} & \textbf{80.52} & \textbf{80.52} & 82.01 \\
			A-comp & 83.75 & 84.56 & \textbf{84.86} & 81.99 & 82.03 & \textbf{84.24} & 81.88 & 82.36 & \textbf{84.66} \\ \hline
		\end{tabular}
		\caption{Comparison under three different distillation strategies.}
		\label{tab:DistillingStrategies}
	\end{table*}

	\section{Experiments}
	
	To evaluate the effectiveness of the proposed framework, we conduct extensive experiments to demonstrate the improvements of graph convolutional networks after introducing the multi-task distillation. 
	\subsection{Experimental Settings}
	\subsubsection{Datasets.} We use four public benchmark datasets for experiments. Cora, Citeseer and Pubmed~\cite{Sen2008cc} are citation networks, where nodes represent papers and edges represent their citations. A-Comp~\cite{Yang2021CPF} is a co-purchase graph extracted from Amazon, where nodes represent products, edges represent the co-purchased relations of products, and features are bag-of-words vectors extracted from product reviews. 
	For three citation datasets, we follow the public split with fixed 20 nodes per class in the training set. For A-Comp, 20 nodes are randomly sampled from each class for training, 30 nodes for validation, and the rest for test.
	
	\subsubsection{Parameter Settings.} The experimental platform is Intel Core i7-8700K, 3.70GHz CPU, NVIDIA GeForce GTX 2080Ti GPU. We randomly initialize the parameters and employ early stopping with a patience of 50 epochs. Adam optimizer is used for optimization with default settings. We set the initial learning rate as 0.01 with weight decay 0.001. The balance hyper-parameter $\alpha$ for the teacher network is tuned and set as $0.1$, and the balance hyper-parameters $\beta_1$ and $\beta_2$ are set as $0.6$ and $0.3$ respectively. 
	
	\subsection{Case Study}
	
	We first carry out a case study to show how self-distillation from multi-tasks resolves the mismatches between node connections and labels. We run GCN and SDSS-GCN on Cora and show how they perform in some selected nodes that are connected by edges but with different labels. 
	
\begin{figure}[htbp]
		\centering
		\subfigure[Ground truth]{\includegraphics[width=0.15\textwidth]{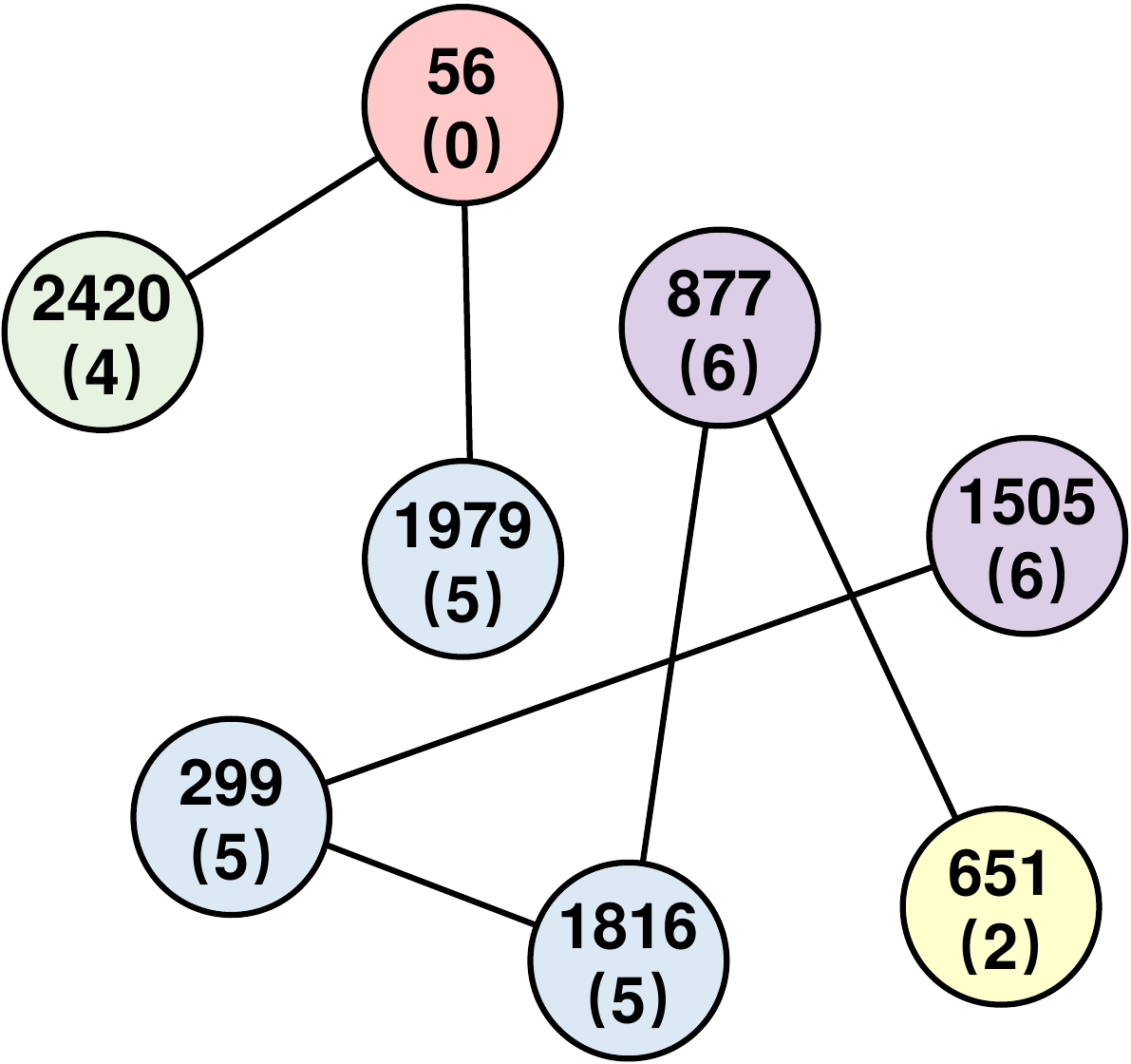}}\label{mismatchGroudTruth}\hspace{0.04in}
		\subfigure[GCN]{\includegraphics[width=0.15\textwidth]{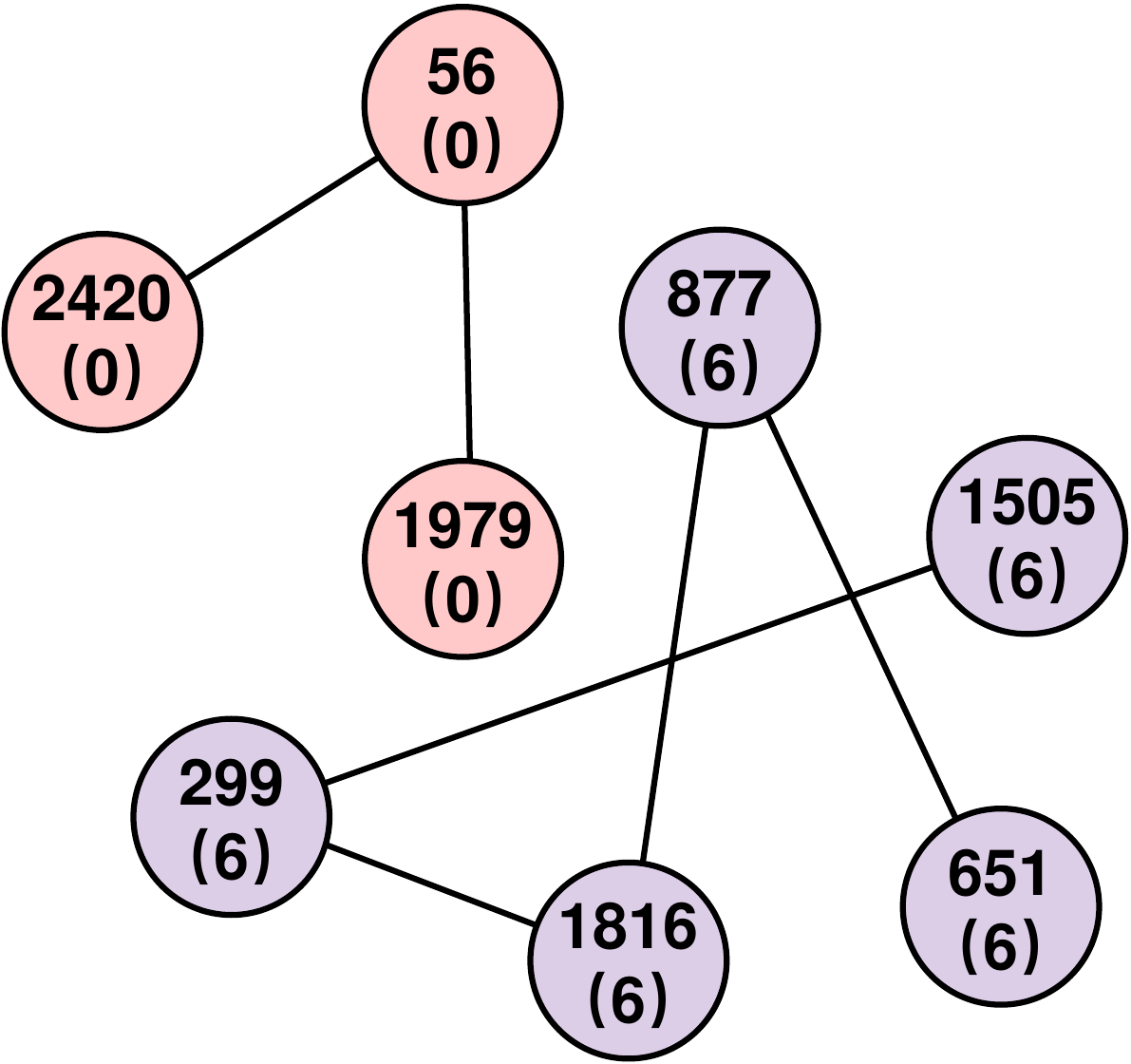}}\label{mismatchGCN}\hspace{0.04in}
		\subfigure[SDSS-GCN]{\includegraphics[width=0.15\textwidth]{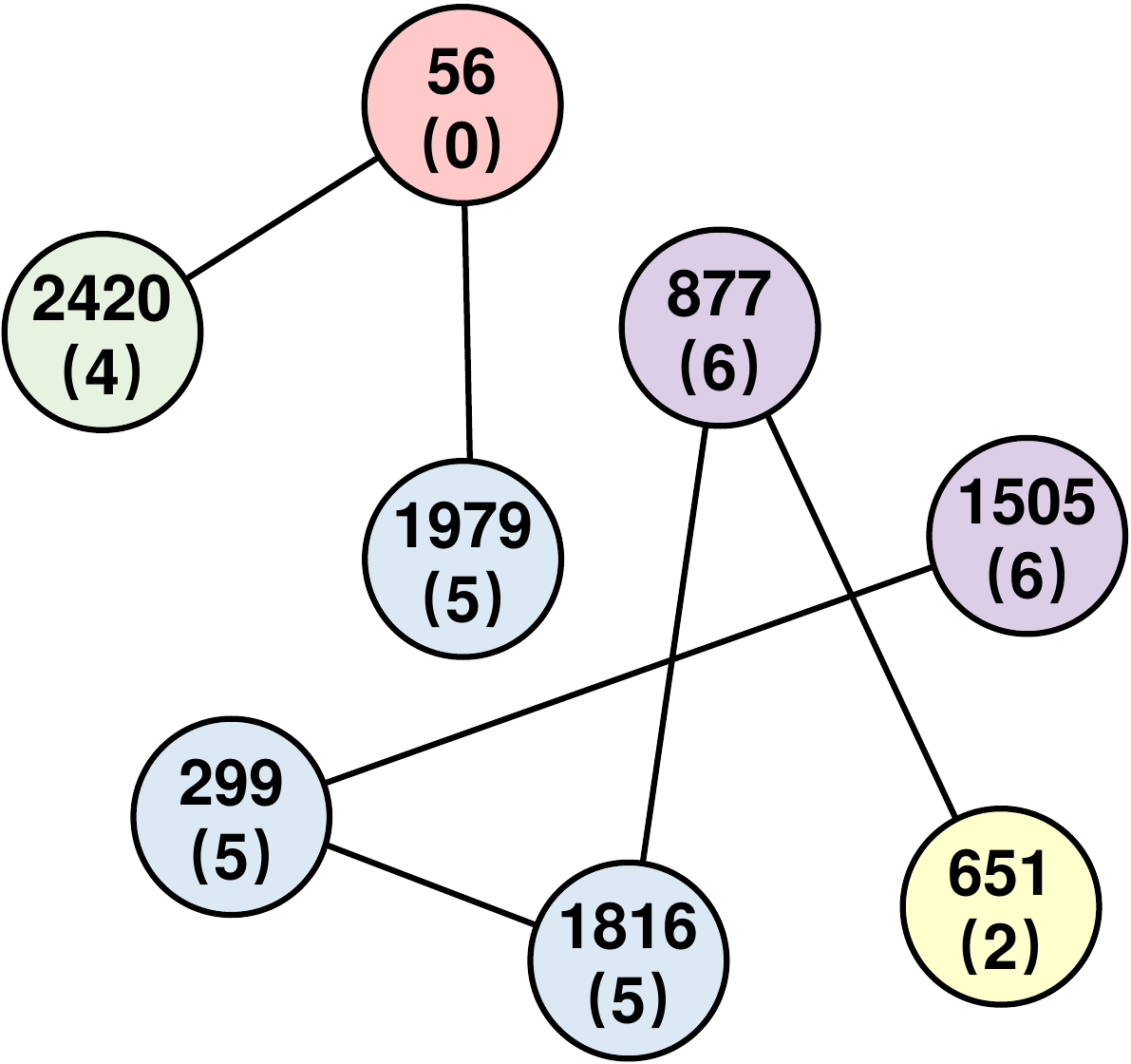}}\label{mismatchSDSS-GCN}
		\caption{Case study: the ground truth, the predictions of GCN and SDSS-GCN on Cora.}
		\label{fig:casestudy}
	\end{figure}

	As shown in Figure~\ref{fig:casestudy}, the numbers within the nodes denote the node indices and their labels. For example, 2420(4) indicates that node 2420 belongs to class 4. The nodes are also colored to show their categories. After feature smoothing, GCN erroneously categories node 2420 and node 1979 into class 0 that is identical to their connected neighbor node 56. Similarly, node 299, node 1816 and node 651 are also put into class 6 by mistake. The results show that the original GCN cannot discriminate whether the labels are coinciding with connections. On the contrary, when introducing multi-task self-distillation, SDSS-GCN can classify nodes 2420, 1979, 299, 1816 and 651 into the correct categories.
	
	\subsection{Ablation Study}
	
	Since our framework is teacher-free and can be integrated into most GCN-based structures, we conduct the ablation study in this subsection to show the roles of different components in our framework. We gradually show that the two strategies used in our paper can largely improve the performance of these frameworks. We select five popular graph convolutional networks for comparison: vanilla GCN~\cite{kipf2016semiGCN}, GraphSAGE~\cite{Hamilton2017GraphSAGE}, GAT~\cite{velivckovic2017GAT}, APPNP~\cite{Klicpera2019APPNP} and GCNII~\cite{chen2020GCNII}.
	
	The results of vanilla GCN, GAT and GraphSAGE are reported in Table~\ref{tab:res1} while results of GCNII and APPNP are presented in Appendix. +SD, +SS, +SDSS denote the corresponding models with self-distillation, with self-supervision, and with both strategies. It can be observed that the two strategies can largely improve the performance of original graph convolutional networks. Besides, the combination of self-distilling and self-supervision brings surprising performance gains. From the data perspective, the average improvements on Cora, Citeseer, Pubmed, A-Comp are 3.70\%, 7.00\%, 5.32\%, 3.96\%, among which Citeseer benefits most from our framework. From the model perspective, the average improvements for GCN, GAT and GraphSAGE are 5.55\%, 4.85\%, 7.77\% respectively. The results are in accordance with our expectations. The improvements mainly come from our incorporating of information from both the data and model sides, which provides additional supervision that can fully utilize the graph structures and limited labels.
	
	\subsection{Comparison with the State-of-the-arts}
	
	In this subsection, we firstly give the overall comparison between our method and 12 graph neural networks on Cora, Citeseer, Pubmed and A-Comp. The baseline models are: GCN~\cite{kipf2016semiGCN}, GAT~\cite{velivckovic2017GAT}, GraphSAGE~\cite{Hamilton2017GraphSAGE}, GCNII~\cite{chen2020GCNII}, APPNP~\cite{Klicpera2019APPNP}, SGC~\cite{wu2019SGCNs}, DropEdge~\cite{rong2020dropedge}, FastGCN~\cite{chen2018fastgcn}, MixHop~\cite{abuelhaija2019mixhop}, Graph U-Net~\cite{gao2019graphUNets}, GMNN~\cite{qu2019gmnn}, and GRAND~\cite{feng2020grand}.
	We also provide the variations of our model based on different teacher networks. We report the results in Table \ref{tab:Sota}. It can be observed that our framework achieves the highest performance compared with baseline methods. Since our method trains the model in two stages, the knowledge from the first stage can be transferred to guide the training of the second stage. The multi-task training in each stage further mines the information within the data, which largely boosts the performance of GCN-based methods. It is surprising to find that simply grafting self-distillation and self-supervision in vanilla GCN can obtain impressive results that are even superior or competitive to advanced graph neural networks. The observations show that simple models may benefit more from the data and model knowledge.

	\begin{figure}[!h]
		\centering
		\subfigure[GCN]{\includegraphics[width=0.23\textwidth]{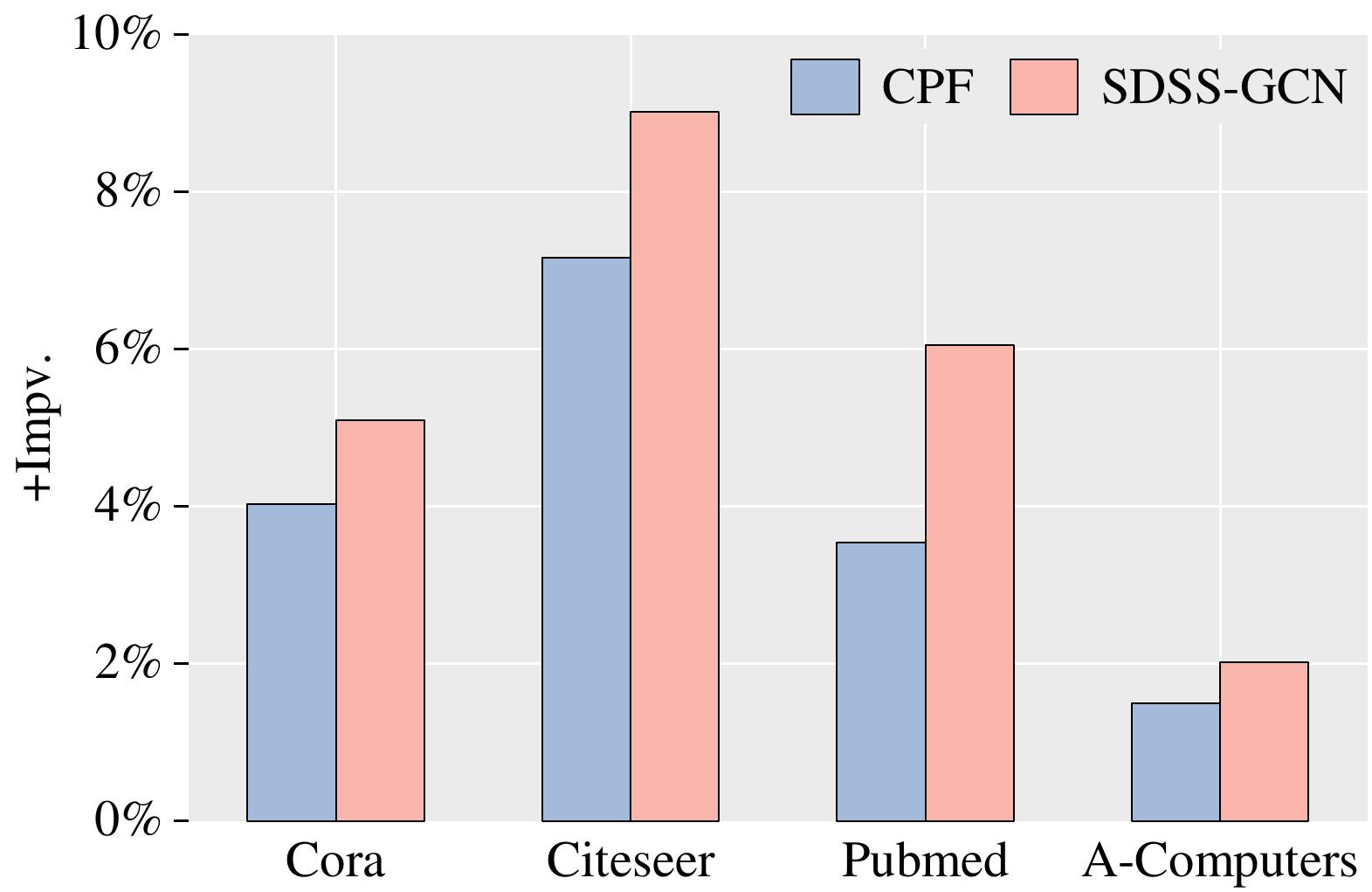}}\label{ImpvGCN}
		\subfigure[GraphSAGE]{\includegraphics[width=0.23\textwidth]{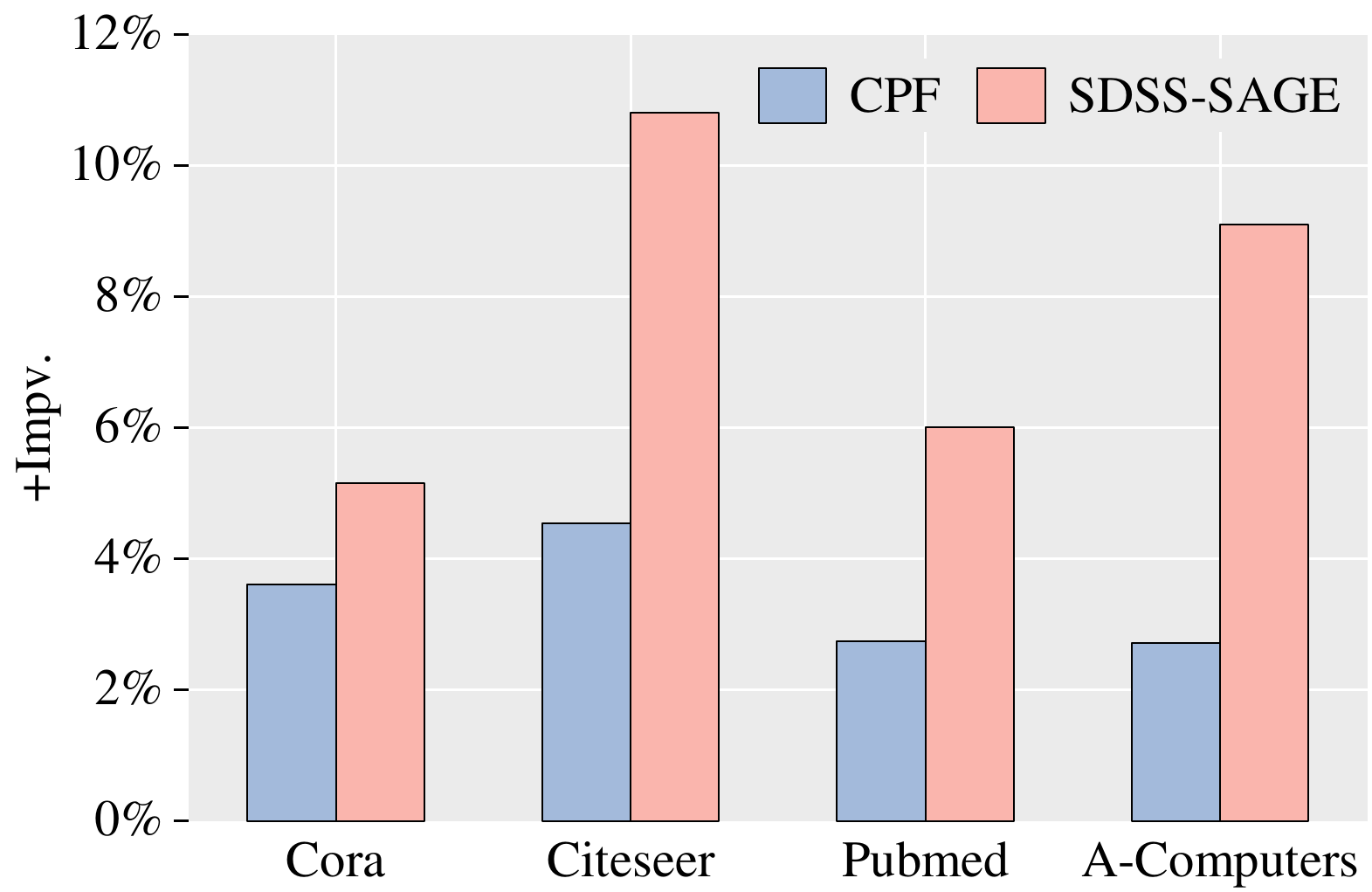}}\label{ImpSAGE}
		\caption{Comparison between SDSS-GCN and CPF in terms of performance improvements for GCN and GraphSAGE.}
		\label{fig:Impv}
	\end{figure}
	
	Secondly, we specially compare our framework with a recent work CPF~\cite{Yang2021CPF} which addresses the semi-supervised learning on graphs via knowledge distilling and label propagation. For simplicity, we report the improvements of our method (SDSS-) and CPF over five graph convolutional networks in Figure~\ref{fig:Impv}. The results for GAT, GCNII and APPNP are listed in Appendix. It shows that the combination of self-supervision and self-distillation achieves maximally 6.38\% higher improvements compared with CPF. In addition, our framework is teacher-free and is applicable to most graph convolutional architectures.
	
	\subsection{Effects of Self-supervised Tasks}
	
	In this subsection, we further explore the effects of different self-supervised tasks on classification performance: predicting node degrees (Deg.), node clustering (Clu.), graph partitioning (Part.), and graph completion (Comp.). As shown in Table~\ref{tab:task1}, the effects of self-supervised tasks are distinct for different models and datasets. For example, In Citeseer, graph partitioning is more suitable for GraphSAGE, and predicting node degrees is more suitable for vanilla GCN. For larger graphs (e.g., Pubmed), node clustering is less competitive compared with other tasks. Graph partitioning is synchronized with GAT since they both focus on detecting the information from graph structures. Besides, the types of self-supervised tasks can also affect the performance, e.g., the regression tasks (Deg. and Part.) are more suitable for GraphSAGE. 
	
	\subsection{Effects of Distillation Strategies.}
	
	In this subsection, we illustrate the effects of three kinds of knowledge distilled during the training: the knowledge from node classification (NC), the knowledge from node classification and middle-layer (NC+M), and the knowledge from node classification, middle-layer, and self-supervised learning (NC+SS+M). As shown in Table~\ref{tab:DistillingStrategies}, adding each kind of knowledge can gradually improve the performance. For GAT and GraphSAGE, the improvements of adding middle-layer information are marginal. For all methods and datasets, adding distillation and self-supervision can boost the performance.
	
	\begin{figure}[htbp]
		\centering
		\includegraphics[width=0.4\textwidth]{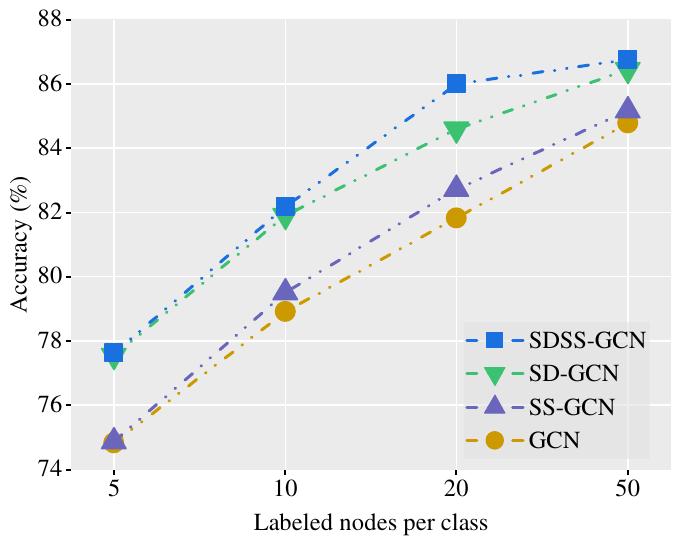}
		\label{layer2}\hspace{0.1in}
		\caption{Classification results under different ratios of labeled nodes on Cora.}
		\label{fig:layerratio}
	\end{figure}
	
	\subsection{Effects of Training Ratios.}
	
	In this subsection, we analyze how SDSS strategy performs under different training ratios. Vanilla GCN is selected as the basic model, and the number of labeled nodes per class varies from 5 to 50. The results of different strategies (GCN, SS-GCN, SD-GCN, SDSS-GCN) on Cora are plotted in Figure~\ref{fig:layerratio}. Under different training ratios, SDSS-GCN always achieves the best performance. Compared with self-supervision, self-distillation is more effective in boosting classification performance. Our method also performs well with few labeled nodes. As evidence, the performance improvements of SDSS-GCN are 4.89\%, 4.14\%, 6.67\% and 3.22\% on average for 5, 10, 20 and 50 labeled nodes per class. Hence, our model is able to further mine the information in graph-based semi-supervised learning.
	
	\section{Conclusion}
	In this paper, we studied the challenge of semi-supervised learning on graphs, and pointed out that using current graph structures is inadequate to capture the class similarities. The labels are shifted and can not be exactly revealed by node connections. From this perspective, we proposed a multi-task self-distillation framework with a two-stage training to combine self-supervision and self-distillation. Compared with current graph-based semi-supervised learning methods, especially the knowledge distilling frameworks, we have the following two main improvements: First, we aimed at distilling knowledge from multi-tasks so that different levels of graph similarities are extracted and can be used to improve the feature aggregation in GCNs. The node features and labels can be aligned with the assist of multi-task distillation. Second, our framework is teacher-free, which relies on no specific GCN structures. As examples, we integrated our framework into several widely-used GCN structures and achieved impressive performance gains practically.

	\bibliography{mybib}

\begin{thebibliography}{34}
\providecommand{\natexlab}[1]{#1}

\bibitem[{Abu-El-Haija et~al.(2019)Abu-El-Haija, Perozzi, Kapoor, Alipourfard,
  Lerman, Harutyunyan, Ver~Steeg, and Galstyan}]{abuelhaija2019mixhop}
Abu-El-Haija, S.; Perozzi, B.; Kapoor, A.; Alipourfard, N.; Lerman, K.;
  Harutyunyan, H.; Ver~Steeg, G.; and Galstyan, A. 2019.
\newblock Mixhop: Higher-order graph convolutional architectures via sparsified
  neighborhood mixing.
\newblock In \emph{ICML}, 21--29.

\bibitem[{Belkin, Niyogi, and Sindhwani(2006)}]{belkin2006manifold}
Belkin, M.; Niyogi, P.; and Sindhwani, V. 2006.
\newblock Manifold Regularization: A Geometric Framework for Learning from
  Labeled and Unlabeled Examples.
\newblock \emph{Journal of Machine Learning Research}, 7(85): 2399--2434.

\bibitem[{Chen, Ma, and Xiao(2018)}]{chen2018fastgcn}
Chen, J.; Ma, T.; and Xiao, C. 2018.
\newblock FastGCN: Fast Learning with Graph Convolutional Networks via
  Importance Sampling.
\newblock In \emph{ICLR}.

\bibitem[{Chen et~al.(2020)Chen, Wei, Huang, Ding, and Li}]{chen2020GCNII}
Chen, M.; Wei, Z.; Huang, Z.; Ding, B.; and Li, Y. 2020.
\newblock Simple and deep graph convolutional networks.
\newblock In \emph{ICML}, 1725--1735.

\bibitem[{Chien et~al.(2021)Chien, Peng, Li, and
  Milenkovic}]{chien2020adaptive}
Chien, E.; Peng, J.; Li, P.; and Milenkovic, O. 2021.
\newblock Adaptive Universal Generalized PageRank Graph Neural Network.
\newblock In \emph{ICLR}.

\bibitem[{Doostparast~Torshizi and Petzold(2018)}]{doostparast2018graph}
Doostparast~Torshizi, A.; and Petzold, L.~R. 2018.
\newblock Graph-based semi-supervised learning with genomic data integration
  using condition-responsive genes applied to phenotype classification.
\newblock \emph{Journal of the American Medical Informatics Association},
  25(1): 99--108.

\bibitem[{Feng et~al.(2020)Feng, Zhang, Dong, Han, Luan, Xu, Yang, Kharlamov,
  and Tang}]{feng2020grand}
Feng, W.; Zhang, J.; Dong, Y.; Han, Y.; Luan, H.; Xu, Q.; Yang, Q.; Kharlamov,
  E.; and Tang, J. 2020.
\newblock Graph Random Neural Networks for Semi-Supervised Learning on Graphs.
\newblock In \emph{NeurIPS}, volume~33, 22092--22103.

\bibitem[{Gao and Ji(2019)}]{gao2019graphUNets}
Gao, H.; and Ji, S. 2019.
\newblock Graph U-Nets.
\newblock In \emph{ICML}, 2083--2092.

\bibitem[{Geoffrey, Oriol, and Jeffrey(2015)}]{hinton2015distilling}
Geoffrey, E.~H.; Oriol, V.; and Jeffrey, D. 2015.
\newblock Distilling the Knowledge in a Neural Network.
\newblock \emph{CoRR}, abs/1503.02531.

\bibitem[{Hamilton, Ying, and Leskovec(2017)}]{Hamilton2017GraphSAGE}
Hamilton, W.~L.; Ying, R.; and Leskovec, J. 2017.
\newblock Inductive Representation Learning on Large Graphs.
\newblock In \emph{NeurIPS}, 1025–1035.

\bibitem[{Hongwei and Jure(2020)}]{wang2020unifying}
Hongwei, W.; and Jure, L. 2020.
\newblock Unifying Graph Convolutional Neural Networks and Label Propagation.
\newblock \emph{CoRR}, abs/2002.06755.

\bibitem[{Karypis and Kumar(1998)}]{Karypis1998Partitioning}
Karypis, G.; and Kumar, V. 1998.
\newblock A Fast and High Quality Multilevel Scheme for Partitioning Irregular
  Graphs.
\newblock \emph{SIAM Journal on Scientific Computing}, 20(1): 359--392.

\bibitem[{Ke, Zhouchen, and Zhanxing(2020)}]{Zhu2020MultiStageSS}
Ke, S.; Zhouchen, L.; and Zhanxing, Z. 2020.
\newblock Multi-Stage Self-Supervised Learning for Graph Convolutional Networks
  on Graphs with Few Labeled Nodes.
\newblock In \emph{AAAI}, 5892--5899.

\bibitem[{Kipf and Welling(2017)}]{kipf2016semiGCN}
Kipf, T.~N.; and Welling, M. 2017.
\newblock Semi-supervised classification with graph convolutional networks.
\newblock In \emph{ICLR}.

\bibitem[{Klicpera, Bojchevski, and G{\"u}nnemann(2019)}]{Klicpera2019APPNP}
Klicpera, J.; Bojchevski, A.; and G{\"u}nnemann, S. 2019.
\newblock Predict then Propagate: Graph Neural Networks meet Personalized
  PageRank.
\newblock In \emph{ICLR}.

\bibitem[{Li, Han, and Wu(2018)}]{li2018deeper}
Li, Q.; Han, Z.; and Wu, X.-M. 2018.
\newblock Deeper insights into graph convolutional networks for semi-supervised
  learning.
\newblock In \emph{AAAI}, 3538--3545.

\bibitem[{Li et~al.(2019)Li, Wu, Liu, Zhang, and Guan}]{li2019label}
Li, Q.; Wu, X.-M.; Liu, H.; Zhang, X.; and Guan, Z. 2019.
\newblock Label efficient semi-supervised learning via graph filtering.
\newblock In \emph{CVPR}, 9582--9591.

\bibitem[{Li, Li, and Wang(2020)}]{li2020co}
Li, S.; Li, W.-T.; and Wang, W. 2020.
\newblock Co-gcn for multi-view semi-supervised learning.
\newblock In \emph{AAAI}, volume~34, 4691--4698.

\bibitem[{Qu, Bengio, and Tang(2019)}]{qu2019gmnn}
Qu, M.; Bengio, Y.; and Tang, J. 2019.
\newblock Gmnn: Graph markov neural networks.
\newblock In \emph{ICML}, 5241--5250.

\bibitem[{Rong et~al.(2019)Rong, Huang, Xu, and Huang}]{rong2020dropedge}
Rong, Y.; Huang, W.; Xu, T.; and Huang, J. 2019.
\newblock DropEdge: Towards Deep Graph Convolutional Networks on Node
  Classification.
\newblock In \emph{ICLR}.

\bibitem[{Sen et~al.(2008)Sen, Namata, Bilgic, Getoor, Galligher, and
  Eliassi-Rad}]{Sen2008cc}
Sen, P.; Namata, G.; Bilgic, M.; Getoor, L.; Galligher, B.; and Eliassi-Rad, T.
  2008.
\newblock Collective Classification in Network Data.
\newblock \emph{AI Magazine}, 29(3): 93.

\bibitem[{Shao et~al.(2018)Shao, Sang, Gao, and Ma}]{shao2018spatial}
Shao, Y.; Sang, N.; Gao, C.; and Ma, L. 2018.
\newblock Spatial and class structure regularized sparse representation graph
  for semi-supervised hyperspectral image classification.
\newblock \emph{Pattern Recognition}, 81: 81--94.

\bibitem[{Subramanya, Petrov, and Pereira(2010)}]{subramanya2010efficient}
Subramanya, A.; Petrov, S.; and Pereira, F. 2010.
\newblock Efficient graph-based semi-supervised learning of structured tagging
  models.
\newblock In \emph{EMNLP}, 167--176.

\bibitem[{Veli{\v{c}}kovi{\'{c}} et~al.(2018)Veli{\v{c}}kovi{\'{c}}, Cucurull,
  Casanova, Romero, Li{\`{o}}, and Bengio}]{velivckovic2017GAT}
Veli{\v{c}}kovi{\'{c}}, P.; Cucurull, G.; Casanova, A.; Romero, A.; Li{\`{o}},
  P.; and Bengio, Y. 2018.
\newblock {Graph Attention Networks}.
\newblock In \emph{ICLR}.

\bibitem[{Wan et~al.(2021)Wan, Pan, Yang, and Gong}]{wan2021contrastive}
Wan, S.; Pan, S.; Yang, J.; and Gong, C. 2021.
\newblock Contrastive and Generative Graph Convolutional Networks for
  Graph-based Semi-Supervised Learning.
\newblock In \emph{AAAI}, volume~35, 10049--10057.

\bibitem[{Wu et~al.(2019)Wu, Souza, Zhang, Fifty, Yu, and
  Weinberger}]{wu2019SGCNs}
Wu, F.; Souza, A.; Zhang, T.; Fifty, C.; Yu, T.; and Weinberger, K. 2019.
\newblock Simplifying Graph Convolutional Networks.
\newblock In \emph{ICML}, 6861--6871.

\bibitem[{Yamaguchi and Hayashi(2017)}]{Yuto2017LPfail}
Yamaguchi, Y.; and Hayashi, K. 2017.
\newblock When Does Label Propagation Fail? A View from a Network Generative
  Model.
\newblock In \emph{IJCAI}, 3224--3230.

\bibitem[{Yang et~al.(2017)Yang, Bai, Zhang, Yuan, and Han}]{yang2017bridging}
Yang, C.; Bai, L.; Zhang, C.; Yuan, Q.; and Han, J. 2017.
\newblock Bridging collaborative filtering and semi-supervised learning: a
  neural approach for poi recommendation.
\newblock In \emph{KDD}, 1245--1254.

\bibitem[{Yang, Liu, and Shi(2021)}]{Yang2021CPF}
Yang, C.; Liu, J.; and Shi, C. 2021.
\newblock Extract the Knowledge of Graph Neural Networks and Go Beyond it: An
  Effective Knowledge Distillation Framework.
\newblock In \emph{The Web Conference}, 1227--1237.

\bibitem[{Yang, Cohen, and Salakhutdinov(2016)}]{Yang2016Revisiting}
Yang, Z.; Cohen, W.~W.; and Salakhutdinov, R. 2016.
\newblock Revisiting Semi-Supervised Learning with Graph Embeddings.
\newblock In \emph{ICML}, 40–48.

\bibitem[{You et~al.(2020)You, Chen, Wang, and Shen}]{Yuning2020Whendoes}
You, Y.; Chen, T.; Wang, Z.; and Shen, Y. 2020.
\newblock When Does Self-Supervision Help Graph Convolutional Networks?
\newblock In \emph{ICML}, volume 119, 10871--10880.

\bibitem[{Zhang and Sabuncu(2020)}]{zhang2020self}
Zhang, Z.; and Sabuncu, M. 2020.
\newblock Self-Distillation as Instance-Specific Label Smoothing.
\newblock In \emph{NeurIPS}, volume~33, 2184--2195.

\bibitem[{Zhou et~al.(2004)Zhou, Bousquet, Lal, Weston, and
  Sch{\"o}lkopf}]{zhou2004learning}
Zhou, D.; Bousquet, O.; Lal, T.~N.; Weston, J.; and Sch{\"o}lkopf, B. 2004.
\newblock Learning with local and global consistency.
\newblock In \emph{NeurIPS}, 321--328.

\bibitem[{Zhu, Lafferty, and Rosenfeld(2005)}]{zhu2005semiGraph}
Zhu, X.; Lafferty, J.; and Rosenfeld, R. 2005.
\newblock \emph{Semi-Supervised Learning with Graphs}.
\newblock Ph.D. thesis, Carnegie Mellon University.

\end{thebibliography}
	
\end{document}